\documentclass{article}
\usepackage{ijcai21}

\usepackage{times}
\usepackage{soul}
\usepackage{url}
\usepackage[hidelinks]{hyperref}
\usepackage[utf8]{inputenc}
\usepackage[small]{caption}
\usepackage{graphicx}
\usepackage{amsmath}
\usepackage{amsthm}
\usepackage{booktabs}
\usepackage{algorithm}
\usepackage{algorithmic}
\usepackage{float}
\usepackage{xcolor,colortbl}
\usepackage{multirow}
\usepackage{amssymb}
\usepackage{mathrsfs}

\usepackage{cellspace}%

\def\eg{\emph{e.g.}}
\def\ie{\emph{i.e.}}

\def\etc{\emph{etc}}

\title{PoseDet: Fast Multi-Person Pose Estimation Using Pose Embedding}

\author{
   Chenyu Tian$^1$
\and
   Ran Yu$^1$\and
   Xinyuan Zhao$^2$\and
   Weihao Xia$^1$\and
   Haoqian Wang$^1$\And
   Yujiu Yang$^1$
\affiliations
$^1$Tsinghua University\\
$^2$Northwestern University\\
\emails
tcy19@mails.tsinghue.edu.cn,
yang.yujiu@sz.tsinghua.edu.cn
}

\begin{document}

\maketitle

\begin{abstract}
     Current methods of multi-person pose estimation typically treat the {\it localization} and the {\it association} of body joints separately. %
    It is convenient but inefficient, leading to additional computation and a waste of time. 
    This paper, however, presents a novel framework PoseDet (Estimating \textbf{Pose} by \textbf{Det}ection) to localize and associate body joints simultaneously at higher inference speed. 
    Moreover, we propose the {\it keypoint-aware pose embedding} to represent an object in terms of the locations of its keypoints.
    The proposed pose embedding contains semantic and geometric information, allowing us to access discriminative and informative features efficiently.
    It is utilized for candidate classification and body joint localization in PoseDet, leading to robust predictions of various poses.
    This simple framework achieves an unprecedented speed and a competitive accuracy on the COCO benchmark compared with state-of-the-art methods. 
    Extensive experiments on the CrowdPose benchmark show the robustness in the crowd scenes.
    Source code is available.

%Code has been made available at: xxx

\end{abstract}

\section{Introduction}
\label{sec:introduction}
Multi-person pose estimation is a fundamental yet challenging task in computer vision, which plays a vital role in human understanding. It has broad applications such as human-robot interaction, virtual reality, sports analysis, \etc., in which real-time performance is essential. The challenge remains that estimating accurate poses of multi-person in the 2D image at a real-time speed.

\begin{figure}
\centering
\includegraphics[scale=0.45]{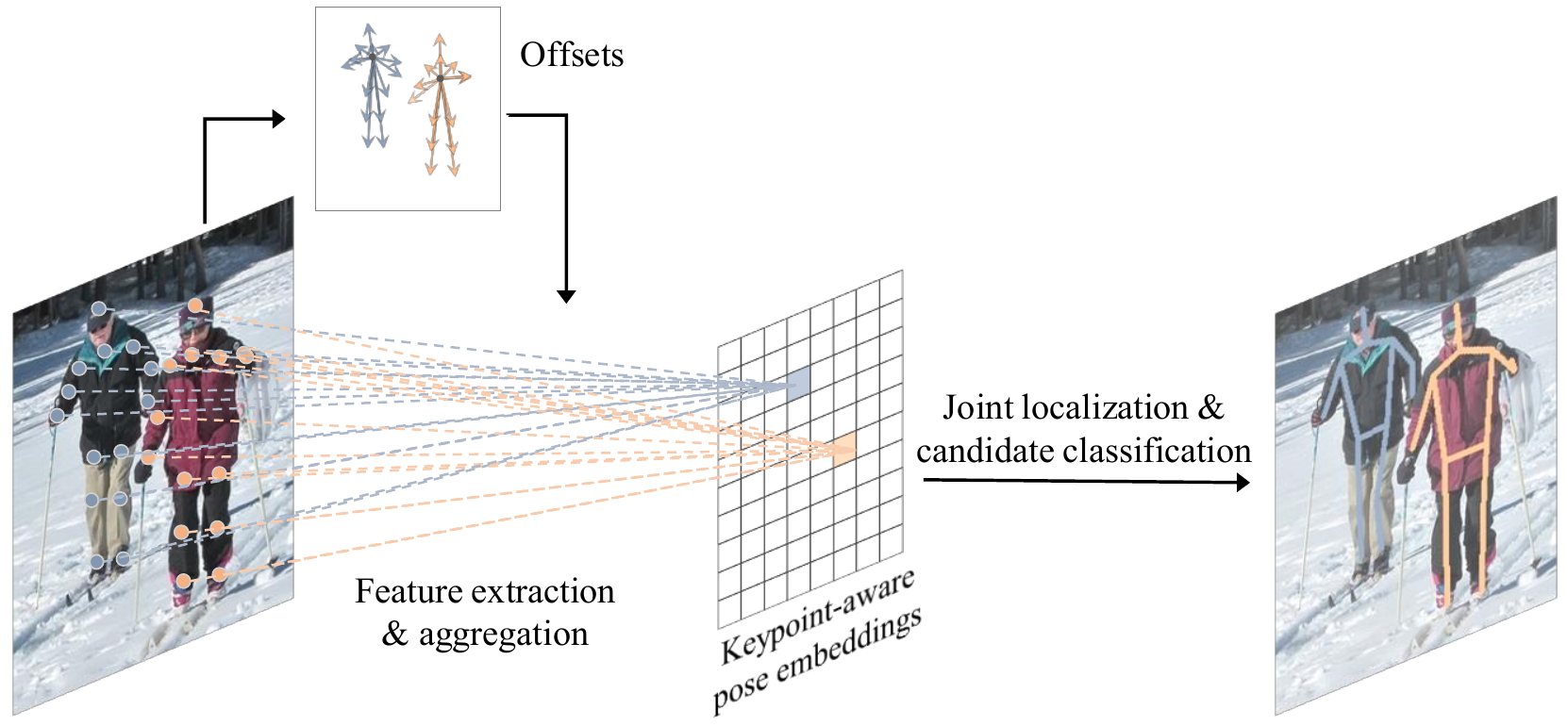}
\caption{Keypoint-aware pose embedding}
\label{fig:pose_embedding}
\end{figure}

As an extension of single-person pose estimation, multi-person pose estimation requires locating all joints and associating the joints with the same person. Current approaches can be categorized as either top-down~\cite{sun2019deep,xiao2018simple,chen2018cascaded,huang2017coarse,papandreou2017towards,fang2017rmpe,guo2019adaptive,ruan2019poinet,jiang2020pay} or bottom-up~\cite{newell2017associative,cao2017realtime,cao2019openpose,Cheng_2020_CVPR,pishchulin2016deepcut,insafutdinov2016deepercut,iqbal2016multi}. Top-down approaches accomplish body joint association by a well-trained person detector. They usually detect person instances by using a person detector, then estimate the single-person pose within the detected box. Bottom-up approaches instead detect all body joints in an image and then associate them with individuals. Both of them carry out a two-stage strategy, which remains an obstacle to real-time inference. 

In this work, we propose a single-stage framework, PoseDet. It combines joint localization and association using a simply designed network to eliminate redundant operations. 
The central idea of PoseDet is to view multi-person pose estimation as an object detection problem, \ie, estimating poses by detection. 
Object detection aims to regress two corner points of the bounding box, while pose estimation can be regarded as the regression of a set of body joints. 
By replacing corner points with key points of body joints, poses can be estimated using an object detection system directly.
However, only the replacement of regression target decreases the performance due to the complexity of regression space. 
Techniques including displacement refinement and keypoint-based operations are adopted, making the detection model more accessible to pose estimation.

Another challenge is the diversity of poses. To accurately regress the keypoints of body joints, the representation of objects is supposed to be informative and fine-grained.
In modern detection systems, the object is usually represented as axis-aligned bounding box~\cite{ren2015faster} or the center point~\cite{tian2019fcos}. 
However, these bounding boxes contained unexpected background areas that might lower the feature quality, and the highly overlapped boxes could be incorrectly suppressed by non-maximum suppression (NMS). 
Representation with the center point faces the problem that center points of distinctive objects could be too close to distinguish. 
Both representations provide rough information of the object and are not suitable for pose estimation. 

A more delicate representation is proposed by RepPoints~\cite{yang2019reppoints}. The object is represented using a set of sample points whose locations are learned in an unsupervised way. Point-set Anchors~\cite{wei2020point} utilize it in pose estimation, where predefined anchors of poses are used as initial locations of sample points. 
Inspired by~\cite{yang2019reppoints,wei2020point}, we propose the keypoint-aware pose embedding. 
As shown in Figure~\ref{fig:pose_embedding}, we use the body joints as keypoints to represent the person instance. 
In conjunction with the deformable convolutional network (DCN)~\cite{dai2017deformable}, we generate the pose embedding by aggregating the extracted feature vectors at locations of keypoints.
Pose embedding is informative as it contains both semantic information of joint category and geometric information of the pose.
The pose embedding directly predicts the refined pose and the score of each instance. 
Different from Point-set Anchors, the locations of keypoints are learnable and supervised by GT keypoints, making them closer to target keypoints during prediction, which allows PoseDet to be robust to various poses. Moreover, PoseDet is an anchor-free detector, leading to lower complexity of the network and a higher inference speed than those of Point-set Anchors.
As a general and flexible representation strategy, the proposed pose embedding can be easily applied to understand other kinds of objects.

Due to the efficient pipeline and the adequate representation, the proposed PoseDet achieves an unprecedented speed and a competitive accuracy for multi-person pose estimation. 
We evaluate our method on COCO~\cite{lin2014microsoft} benchmark. The inference speed of PoseDet is 2.5 times higher (10 FPS) than Point-set Anchors with the same scale of the input image and backbone HRNet-W48~\cite{sun2019deep}. 
By replacing the backbone with DLA-34~\cite{yu2018deep}, the model runs faster at 42 FPS with the accuracy's decline.
The effectiveness of keypoint-aware pose embedding is demonstrated on CrowdPose~\cite{li2019crowdpose} benchmark. PoseDet achieves comparable accuracy with the state-of-the-art solutions and shows its robustness in the crowd scenes.

The contributions of our work are:
\begin{itemize}
\item We reformulate the multi-person pose estimation as an object detection problem. Our proposed PoseDet framework achieves an unprecedented speed and a competitive accuracy compared with state-of-the-art approaches;
\item We propose the keypoint-aware pose embedding as a general and flexible representation for person instances, making PoseDet being robust to real-world crowd scenes with occlusion and various poses. It can be easily applied to other instance-level recognition tasks like person re-identification;
\item The performance of PoseDet is further improved by displacement refinement and keypoint-based operations.
\end{itemize}

\section{Related Work}
Due to the various applications in the real world, human pose estimation is an active research topic for decades. In the early works, joint locations are predicted based on hand-craft features, such as histogram of oriented gradient~\cite{sun2011articulated,wang2013beyond}. Recently, state-of-the-art performance is achieved using the Convolutional Neural Networks (CNNs). In this work, based on CNNs, we focus on simplifying detection procedures and improving object representation for multi-person pose estimation.

\textbf{Real-time multi-person pose estimation.} Multi-person pose estimation requires dense prediction over spatial locations as well as detection of the person instance. The complexity makes it hard to inference at real-time speed. The key is to optimize processing procedures and eliminate redundant structures.~\cite{cao2017realtime} is the first real-time algorithm to tackle this problem. They generate the ground-truth (GT) confidence maps to localize joints and associate them via part affinity fields. ~\cite{kocabas2018multiposenet} predict the heatmap of joints for joint localization and associate them with a person detector and the pose residual network, where all subnets share the same backbone. ~\cite{nie2019single} present SPM, a single-stage solution that predicts hierarchical offsets of joints based on detected root points. The single-stage strategy makes it faster than two-stage (\eg, bottom-up or top-down) strategies. We present a faster single-stage solution that combines joint localization and association in the same pipeline.

\textbf{Object representation in multi-person pose estimation.} Object representation serves as a guide to extract features for purposes such as object classification and location refinement.~\cite{he2017mask} use the bounding box to represent an object. Features are extracted by the bounding box to predict the heatmap of joints. Box-based representation faces the problem of including the non-target area, resulting in low-quality feature and computation overhead.~\cite{zhou2019objects} model an object as a single point and regress joint offsets at each point. Center point-based representation is fast but contains limited information for dense prediction of joints.~\cite{wei2020point} obtains features at a set of predefined anchors to acquire informative features. For refinement of segmentation boundary,~\cite{kirillov2020pointrend} regards an object as a continuous entity, which is encoded in feature maps and can be accessed at any points by interpolation. We follow the point-set representation and the concept of continuous entity, presenting pose embedding that extracts keypoint-aware features.

\section{Approach}

\begin{figure*}
\centering
\includegraphics[scale=0.76]{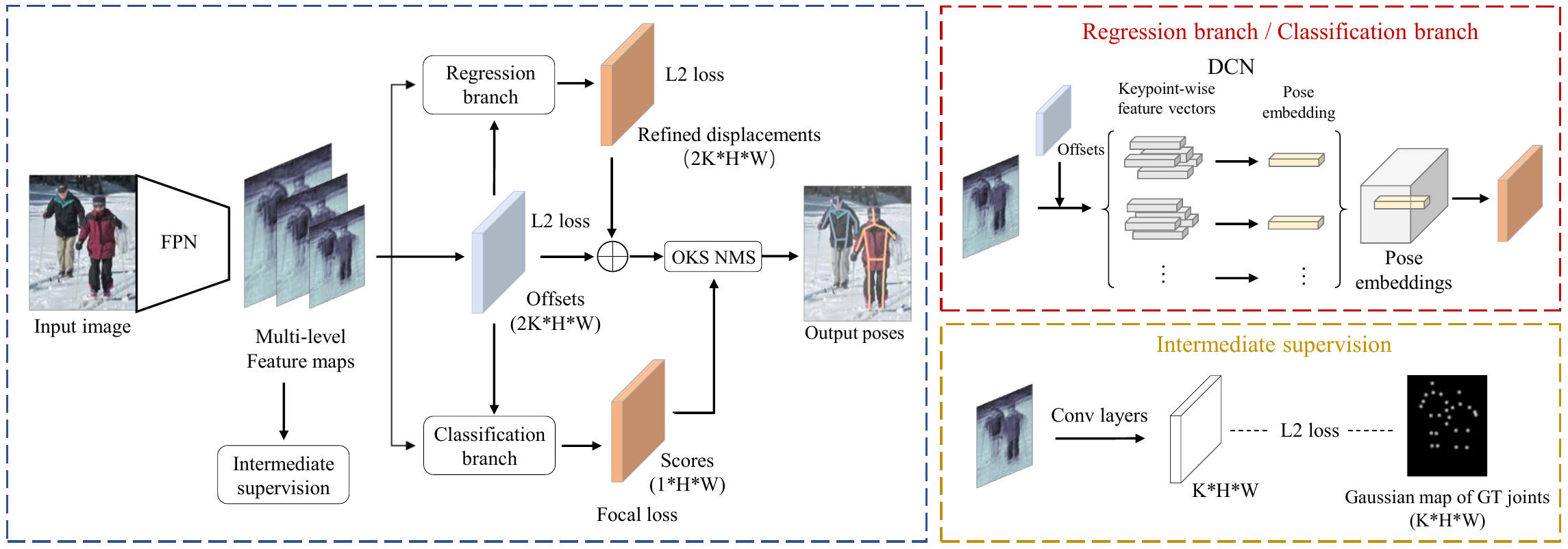}
\caption{An overview of PoseDet. We reformulate multi-person pose estimation as an object detection problem. Regression and classification branch conduct dense predictions over feature maps.}
\label{fig:pipeline}
\end{figure*}

In this section, we first elaborate keypoint-aware pose embedding, a key component of PoseDet in understanding objects effectively and efficiently. Next, we present the multi-level anchor-free detector and show how to utilize pose embedding in it to estimate 2D poses of persons in an image.

\subsection{Keypoint-Aware Pose Embedding}

Object representation is an important medium to recognize objects in visual tasks (\eg, classification and regression). As discussed in Section~\ref{sec:introduction}, features extracted by the bounding box and center point are either redundant or limited for pose estimation. In our system, objects are represented as keypoints, and features at keypoints are extracted and aggregated for classification and regression.

Given the preceding feature maps, the person instance can be regarded as a continuous entity encoded in the feature maps. As shown in Figure~\ref{fig:pose_embedding}, the feature vectors at locations of body joints contain semantic and geometric information of the pose.  

In natural language processing, discrete text data are represented as continuous vectors via word embedding to capture semantic and syntactic features~\cite{hinton1986learning}. Analogously, we extract and aggregate the irregular feature vectors to represent the person instance. The representation is mapped from a set of vectors to one vector, named pose embedding. Formally, the process can be summarized as:
\begin{equation}
f: \{\psi_i \}_{i=1}^K \rightarrow \varepsilon,
\end{equation}
where $\psi_i$ donates feature vector at point $i$ on the feature maps, $K$ is the number of joints. $\varepsilon$ is a vector, named pose embedding, containing the information of $K$ joints. 

The pose embedding contains semantic and geometric information because both the shape and joint categories are confirmed. It's an informative representation that can be used to estimate pose and classify positive or negative candidates, but it's hard to make full use of them by conventional CNNs due to the restriction of the local receptive field and various poses.  
Considering dense hypotheses over spatial locations of feature maps in modern detection networks, which means dense pose embeddings are required. Following~\cite{yang2019reppoints,wei2020point}, we generate dense pose embeddings for these hypotheses via one DCN layer simultaneously. The DCN layer takes offsets as well as feature maps as input to generate pose embeddings. The offset is a 2K-dimensional vector for each hypothesis, which indicates the positions of K keypoints based on the location of the hypothesis on the feature map. DCN generates the pose embedding for each hypothesis according to different offsets, which allows us to access features at arbitrary positions without limitation of geometric structure efficiently. 

The learning of offsets is crucial to the performance of DCN. \cite{dai2017deformable} adopt an unsupervised way to drive the learning of offsets implicitly. \cite{yang2019reppoints} use weak supervision by minimizing the loss between the GT bounding boxes and pseudo boxes generated by offsets. \cite{wei2020point} use the most common poses as predefined anchors to generate fixed offsets. In this work, a bypass is applied to generate the keypoint-aware offsets, which is supervised by GT joints, driving the learning of offsets explicitly. This allows us to generate meaningful feature representation stably. Moreover, dynamic offsets provide more flexibility than\cite{wei2020point}, making our PoseDet robust to various poses.

Pose embedding is applied in PoseDet to estimate pose. As a general representation strategy, we note that it's easy to expand to other objects by replacing the joint with other kinds of keypoint.

\subsection{PoseDet}

Recently, one-stage detectors show compatible accuracy to two-stage strategies with higher inference speed. For fast pose estimation, we start from RepPoints~\cite{yang2019reppoints}, a one-stage anchor-free detector. By utilize pose embedding in RepPoints, we propose PoseDet, a pose estimation machine that combines human detection and poses estimation into the same pipeline.

The pipeline of our approach is illustrated in Figure~\ref{fig:pipeline}. PoseDet comprises a backbone for feature extraction, a bypass for offsets prediction, and two branches for joint localization and candidate scoring. 

The backbone is a feature pyramid network (FPN), producing a multi-scale feature pyramid from a single resolution image. Objects are assigned to different levels according to the scale, which partially solves the problem of scale diversity of poses. PoseDet makes dense hypotheses over the spatial location of multi-level feature maps, also known as candidates. We estimate the pose of each candidate and score it.

Let $\mathbb{F}$ be the single-level feature maps of size  $C \times H \times W $. There are $H \times W $ candidates at that level, processed by the two branches to estimate $H \times W $ possible poses. The bypass produces $H \times W $ sets of offsets, where each set is a 2K-dimensional vector, indicating the offsets from the location of a candidate to the body joints of a person instance. Given the predicted offsets and feature maps extracted by FPN, the other two branches implement task-specific pose embedding for displacement refinement and candidate scoring, respectively. We utilize the above full convolutional networks on each level of FPN, producing dense multi-level predictions, then adopt NMS to eliminate redundant predictions.

In the following sections, we describe the details of regressing body joints and scoring candidates and the intermediate supervision of FPN. Finally, we introduce training and inference of PoseDet, including assignment strategies and NMS.

\subsubsection{Joint Localization}

Multi-person poses in an image can be represented as:

\begin{equation}
\mathbb{P} = \{ (x_1^i, y_1^i), (x_2^i, y_2^i),...,(x_K^i, y_K^i) \}_{i=1}^N,
\end{equation}
where N is the number of persons in images, and $(x_j^i, y_j^i)$ donates the coordinates of $j$ the joints of person $i$. To estimate the pose of a person instance, we regress the displacements from the candidate's point to GT joints (each candidate is assigned with a GT pose). It's hard to regress the displacements directly due to the variety of poses. To tackle the problem, following~\cite{nie2019single}, we implement a coarse-to-fine manner to approximate a more accurate location estimation. The process can be formulated as:

\begin{equation}
(x_j^i, y_j^i) = (x, y) + (\delta_{x}^c, \delta_{y}^c) +  (\delta_{x}^f, \delta_{y}^f).
\label{eq:pose}
\end{equation}
Here, (x, y) is the coordinate of the candidate, $(\delta_{x}^c, \delta_{y}^c)$ is coarse displacement, and $(\delta_{x}^f, \delta_{y}^f)$ is refined displacement.

The offsets predicted by the bypass provide a coarse prediction of the pose, which not only guides DCN to generate pose embedding but also gives the coarse displacements. We estimate the refined displacements based on the coarse prediction. The regression branch is composed of two $3\times 3$ conv layers, and a deformable convolution layer is applied to produce pose embedding. The refined displacements are regressed directly via $1\times 1$ conv layer, which takes pose embedding as input. We get the final prediction by aggregating the coarse prediction and the refinement (Equation \ref{eq:pose}).

Pose embedding encodes the bias of coarse prediction, leading to a more precise estimation, solving the problem of regressing offsets on a large scale, high-dimensional vector space.

\subsubsection{Candidate Classification}

Based on the coarse prediction, we evaluate the candidate by classifying positive and negative predictions, where the predicted score indicates the probability of positive prediction. We use Object Keypoint Similarity (OKS) as a metric to evaluate the similarity between predicted pose and GT pose. Predictions with OKS higher than 0.6 indicate positive, lower than 0.5 indicate negative. OKS is a collection of distances between every two joints of the same category (\ie, the similarity between two point sets). It provides a point-wise evaluation for the evaluation of poses.

The classification branch after FPN is used to generate pose embedding for candidate scoring, which is composed of two $3\times 3$ conv layers and a deformable convolution layer. A following $1\times 1$ conv layer is applied to estimate the score. The predicted scores and poses are used in NMS to suppress redundant predictions.

\subsubsection{Intermediate Supervision}

The feature map of FPN is considered a mapping of the person instance, where we extract the object representation to generate keypoint-aware pose embeddings. To force the FPN to be sensitive to body joints, following~\cite{newell2016stacked,wei2016convolutional}, we utilize intermediate supervision with GT joints. Specifically, we add a branch consisted of a $3\times 3$ conv layer and a $1 \times 1$ conv layer, which takes the multi-level feature maps of FPN as input and generates multi-level heatmaps with the same resolutions. The heatmaps indicate the probabilities of joints over spatial locations, where the number of channels is equal to the number of joints. Naturally, for the heatmaps of each level, we generate 2D Gaussian maps (with a standard deviation of 2 pixels) centered on the GT joint location as targets.  We minimize the $l_2$ loss between predicted heatmaps and targets during training.

We note that the intermediate supervision is a cost-free improvement because this branch is not involved in testing. We demonstrate the benefit of it in ablation study (Section~\ref{sec:ablation}).

\begin{figure}
\centering
\includegraphics[scale=0.57]{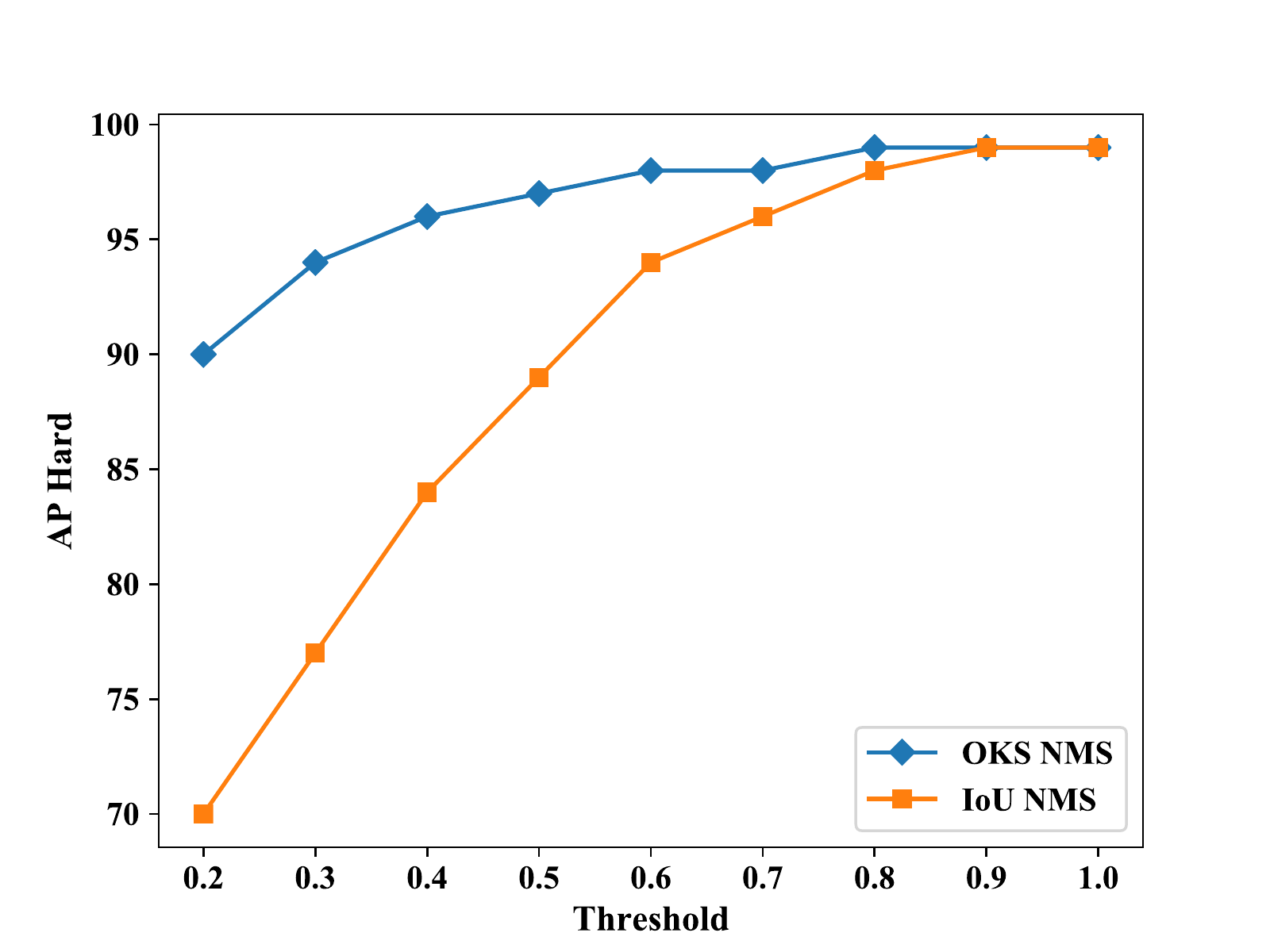}
\caption{Upper bounds of OKS-based NMS and IoU-based NMS on CrowdPose.}
\label{fig:NMS}
\end{figure}

% \vspace{-0.1cm}
\subsubsection{Training and Inference}
\label{sec:nms}
In the training phase, we use different losses and assigning strategies in the bypass, the classification branch, and the regression branch. The loss of network can be summarized as:

\begin{equation}
	\begin{split}
    Loss = L_2(GT Pose, \delta^c) + L_2(GT Pose -  \delta^c,  \delta^f) \\
    + FL(GT, scores) + L_2(GM, Heatmaps)
	\end{split}
\end{equation}

On the right side of the equation, the first item is loss of the bypass. The bypass is responsible for providing a coarse prediction of each candidate. We use GT joints as supervision. Each GT pose is allocated to a level in FPN according to its size and assigned to a candidate at the same level closest to the centroid of the GT pose. The centroid is calculated by mapping the spatial locations of the joints to feature maps and averaging the coordinates. We demonstrate the advantage of the centroid of joints compared with the centroid of the bounding box in Section~\ref{sec:ablation}. Candidates assigned with targets are trained with $l_2$ loss. The second item is the loss of the regression branch, which is responsible for predicting more delicate poses based on the coarse predictions. We compute OKS between the coarse prediction and every GT pose, then assign the larges OKS to the candidate. Candidates with OKS higher than 0.7 are trained with $l_2$ loss. The third item is the loss of the classification branch. Both positive and negative candidates are trained with focal loss~\cite{lin2017focal} (FL) to address the large class imbalance issue. The last item is loss of intermediate supervision, which is $l_2$ loss between Gaussian maps (GM) and predicted heatmaps.

For the inference, we predict each candidate's pose and confidence score, then implement OKS-based NMS~\cite{papandreou2017towards} to remove redundant predictions. 

Most of the modern detection systems use the IoU of bounding boxes as the measurement to evaluate the overlap of predictions. When two predictions have an overlap over the predefined threshold, the prediction with a lower confidence score is suppressed. One problem is that the IoU-based NMS might suppress positive samples that occlude heavily, which is not suitable for multi-person pose estimation, especially for the crowd scenes. 

In our system, the overlap is measured in terms of OKS. OKS is dependent on distances between joints, leading to a more discriminate measurement of the similarity. To examine the upper bound on persons with a high crowd index, we use GT keypoints in the CrowdPose test set as input, performing OKS-based NMS and IoU-based NMS on different thresholds, respectively. AP Hard is adopted as the metric that contains the most crowd. As shown in Figure~\ref{fig:NMS}, the IoU-based NMS wrongly suppress more positive samples than OKS-based NMS, showing that the OKS is a better measurement to distinguish occluded persons.

\section{Experiments}
\definecolor{Gray}{gray}{0.85}
\definecolor{Gray1}{gray}{0.55}
\definecolor{Gray2}{gray}{0.7}
\definecolor{Gray3}{gray}{0.85}
\newcolumntype{a}{>{\columncolor{Gray}}r}
\begin{table*} [!ht]
\centering
%\begin{tabular}{lrcarrrra}
\begin{tabular}{lccrrrrrr}
\toprule
Method  & Backbone  & Input size & $\text{FPS}$ & $\text{AP}$  & $\text{AP}^{\text{50}}$ & $\text{AP}^{\text{75}}$ & $\text{AP}^{\text{M}}$ & $\text{AP}^{\text{L}}$\\
\midrule
Ass.Embedding (2017)     & Hourglass-4    
&512   & 4  &56.6 & 81.8 & 61.8 & 49.8 & 67.0      \\
CMU-Pose (2017)            & 3CM-3PAF(102)         
&368 & \cellcolor{Gray3}10   &61.8 & 84.9 & 67.5 & 57.1 & 68.2      \\
PersonLab (2018)       & ResNet-152         
&1401  & 2  & \cellcolor{Gray2} 66.5 & 88.0 & 72.6 & 62.4 & 72.3     \\
Higher HRNet (2020)  & HRNet-W48
&640  & 4 & 65.7 &86.3  &72.3  &60.8  &72.6     \\   % PoseDet_HRNetw32_2 
Poi. Anchors (2020)   & HRNet-W48          
&640  & 4  &\cellcolor{Gray1}67.4 & 88.3 & 74.6 & 64.7 & 73.0    \\
\midrule
PoseDet (Ours)      & HRNet-W48   %PoseDet_HRNetw48_ml e200
& 640 & \cellcolor{Gray3}10 & \cellcolor{Gray3}66.2 & 87.5 & 73.2 & 62.3 & 73.2 \\
PoseDet (Ours)      & HRNet-W32         
&512  & \cellcolor{Gray2}18 & 64.4  & 86.7 & 71.0 & 58.8 & 73.3    \\   % 
PoseDet (Ours)      & DLA-34            
&512 & \cellcolor{Gray1}42  &59.2 & 84.3 & 64.6 & 52.7 & 68.8    \\ % PoseDet_dla34c128_ml2
\bottomrule
\end{tabular}
\caption{Comparison with state-of-the-art bottom-up approaches on COCO \textit{test-dev}, both keypoints $\text{AP}$ and $\text{FPS}$ are obtained without TTA. Input size indicates the sort side of an input image.}
\label{tab:mainresults}
\end{table*}

\subsection{Experimental Setting}
\textbf{Dataset.} We evaluate PoseDet for accuracy and speed of multi-person pose estimation on a widely adopt benchmark, COCO, and evaluate its robustness in the crowd scenes on the CrowdPose benchmark. 

COCO is a large dataset contains over 200K images, divided into \textit{train}/\textit{val}/\textit{test-dev} splits. Each person is labeled with 17 body joints. Models are trained using \textit{train} split (64k images). We report results on COCO \textit{test-dev} split (20K images) for comparisons to state-of-the-art methods. The official Average Precision (AP), calculated in terms of OKS between the prediction and GT pose, is adopted as a metric to evaluate the accuracy, and Frames per Second ($\text{FPS}$) is adopted for the speed.

CrowdPose consists of 20,000 images, including about 80,000 persons. Each person is labeled with 14 body joints. CrowdPose is divided into train, validation, and test sets in the proportion of 5:1:4. The average intersection over union(IoU) of human bounding boxes is 0.27 (0.06 in COCO). We evaluate PoseDet on CrowdPose to validate the robustness in the crowd scenes. The test set of CrowdPose is divided into three crowding levels, indicated as easy, medium, and hard. Three metrics, $\text{AP}^{\text{E}}$, $\text{AP}^{\text{M}}$, and $\text{AP}^{\text{H}}$ are adopted respectively, where $\text{AP}^{\text{H}}$ contains the most crowd.

\textbf{Implementation Details.} We follow the conventional data augmentation strategies and training schedules for multi-person pose estimation. The COCO and CrowdPose dataset share the same training/testing settings for convenience.

During model training, images are augmented with horizontally flipping and resize factor in $[ 0.5, 1.5 ]$, then cropped centered on a random person to $512 \times 512$ patches. The cropped patches are fed into the network as training samples. All models are optimized with Adam~\cite{kingma2014adam} over 8 NVIDIA V100 GPUs with a mini-batch of 48 patches (6 patches per GPU). The initial learning rate is 0.0001, decreased at epoch 180 and 200 by a factor of 0.1. Training converges at epoch 210. We examine different backbones before FPN, which is initialized with the weights pre-trained on  ImageNet~\cite{deng2009imagenet}. 

For evaluation, we conduct single-scale and multi-scale testing. For single-scale testing, unless specified, the input images are resized to have a shorter side of 640 and a longer side less than 1000, with an unchanged aspect ratio. For multi-scale testing, the input image is resized to an image pyramid with a sorter side of $\{300,400,500,600,700,800\}$ and augmented by horizontal flip. NMS is conducted over the results of augmented images. The inference speed is evaluated on GPU GTX 1080Ti and CPU Intel i9-7980XE. All experiments are implemented based on mmdetection~\cite{mmdetection}.

\subsection{Analysis of Accuracy and Speed}

Both accuracy and speed are important in real-world applications of multi-person pose estimation, while current pose estimation metrics are based on the accuracy of keypoints. To provide a more complete comparison ,we evaluate approaches on COCO \textit{test-dev} and present keypoints $\text{AP}$ and inference speed (FPS) in Table~\ref{tab:mainresults}. We compare the performances of bottom-up approaches without test-time augmentation (TTA).

For multi-person pose estimation, models with the highest accuracy are dominated by top-down approaches~\cite{sun2019deep,cai2020learning}. The bounding boxes of persons are detected by a person detector and resized to a fixed size, which avoids the scale diversity, leading to higher accuracy and lower inference speed than bottom-up approaches. For example, HRNet\cite{sun2019deep} achieves keypoints $\text{AP}$ of 75.5 on COCO \textit{test-dev}. They perform single-person pose estimation on detected boxes (20 boxes per image roughly, each box is resized to $384 \times 288$), with an inference time of about 90 ms per box. The performance of inference speed can not be analyzed due to the inference time of the person detector is not reported. We only consider bottom-up approaches in Table~\ref{tab:mainresults}.

Moreover, TTA, including multi-scale testing and flip, is widely adopted in multi-person pose estimation. 
With TTA, too big or too small targets can be resized to a more proper size, making them easier to be detected. 
TTA can improve the accuracy by additional detection but increases inference time and computation overhead. 
TTA with a six-scale image pyramid and flip increase the computation by about 12 times.
We argue that discarding TTA is more practical in application scenarios where real-time performance is required.
Hence, we provide a consistent comparison in Table~\ref{tab:mainresults}, where both $\text{AP}$ and $\text{FPS}$ are obtained without TTA.
% \vspace{-0.5cm}

Specifically, for CMU-Pose~\cite{cao2017realtime} and Associative Embedding~\cite{newell2017associative}, we report the results cited from their papers or GitHub repositories. For Point Anchors~\cite{wei2020point} and Higher HRNet~\cite{Cheng_2020_CVPR}, the results are based on our implementation, because either $\text{AP}$ or $\text{FPS}$ without TTA is not reported by authors. 

Given the results, we can see that PoseDet outperforms other methods in terms of the tradeoff between accuracy and speed. The inference speed of PoseDet with backbone HRNet-W48 is 2.5 times higher than Point Anchors and Higher HRNet with the same backbone and scale of the input image. PoseDet with a lighter backbone DLA-34 achieves a maximum speed (42 $\text{FPS}$) with lower accuracy (59.2 $\text{AP}$). The main reason why backbone HRNet achieves higher accuracy than DLA, especially refers to $\text{AP}^{\text{75}}$, is that HRNet extracts high-resolution representation rich features, which are significant for the pixel-level prediction of joints. To best our knowledge, PoseDet is the fastest approach for multi-person pose estimation.

Without consideration of the inference speed, we also report results with TTA in Table~\ref{tab:TTA}. The comparison indicates the ability of PoseDet to estimate accurate poses. All reported numbers for comparison are cited from the papers. 

\begin{table}
\centering
\scalebox{0.75}{
    \begin{tabular}{ll|rrrrr}
    \toprule
    Method  & Backbone  & $\text{AP}$ & $\text{AP}^{\text{50}}$ &$\text{AP}^{\text{75}}$ & $\text{AP}^{\text{M}}$ & $\text{AP}^{\text{L}}$ \\
    \midrule
    PersonLab & ResNet-152
    & 68.7 & 89.0 & 75.4 & 64.1 & 75.5  \\
    Ass.Embedding & Hourglass-4    
    & 65.5 & 86.8 & 72.3 & 60.6 & 72.6  \\
    SPM & Hourglass-8    
    & 66.9 & 88.5 & 72.9 & 62.6 & 73.1  \\
    Higher HRNet  & HRNet-W48 
    & 70.5 & 89.3 & 77.2 & 66.6 & 75.8  \\
    Poi. Anchors        & HRNet-W48           
    & 68.7 & 89.9 & 76.3 & 64.8 & 75.3  \\
    PoseDet (Ours)      & HRNet-W48            
    & 67.8 & 89.3 & 75.0 & 63.3 & 75.5  \\   % PoseDet_HRNetw48_ml e200 type3 flip nms300/0.005
    PoseDet (Ours)      & HRNet-W32         
    & 66.7 & 88.5 & 73.7 & 62.2 & 75.5  \\   % PoseDet_HRNetw32_4 type3 flip nms300/0.005
    \bottomrule
    \end{tabular}
    }
\caption{Performance on COCO \textit{test-dev} with TTA.}
\label{tab:TTA}
\end{table}

\subsection{Robustness in the Crowd Scenes}

% \begin{table}[htbp]
\begin{table}
\centering
\scalebox{0.85}{
    \begin{tabular}{l|rrrrrr}
    \toprule
    Method  & $\text{AP}$ & $\text{AP}^{\text{50}}$ & $\text{AP}^{\text{75}}$ & $\text{AP}^{\text{E}}$ & $\text{AP}^{\text{M}}$ & $\text{AP}^{\text{H}}$ \\
    \midrule
    Mask-RCNN       
    &57.2 & 83.5 & 60.3 & 69.4 & 57.9 & 45.8     \\ % 
    AlphaPose  
    &61.0 & 81.3 & 66.0 & 71.2 & 61.4 & 51.1    \\ % 
    Simple Baseline
    &60.8 & 81.4 & 65.7 & 71.4 & 61.2 & 51.2 \\
    SPPE 
    &66.0 & 84.2 & 71.5 & 75.5 & 66.3 & 57.4     \\ %
    Higher HRNet   
    & \textbf{67.6} & 87.4 & 72.6 & \textbf{75.8} & \textbf{68.1} & \textbf{58.9}     \\
    % PoseDet (Ours)       
    % &65.1 & 86.9 & 70.3 & 72.9 & 66.0 & 55.9     \\   % PoseDet_HRNetw48_ml_crowd      
    PoseDet (Ours)        
    &67.3 & \textbf{88.9} & \textbf{72.8} & 75.2 & 68.0 & 58.2    \\
    \bottomrule
    \end{tabular}
}
\caption{Comparison with state-of-the-art methods on CrowdPose with TTA.}
\label{tab:crowdpose}
\end{table}

Another important issue in real-world applications is the robustness to various poses, including occluded persons and various poses. Most of images in popular datasets (\eg, COCO~\cite{li2019crowdpose} and MPII~\cite{andriluka20142d}) have no overlapped persons. We evaluate the robustness of PoseDet on the CrowdPose benchmark, which has a larger crowd index and more persons in an image than COCO. The inference speed of PoseDet is not affected by the number of persons, which is the same on CrowdPose as COCO, so we focus on evaluating the accuracy on CrowdPose. PoseDet with HRNet-W48 backbone is trained using the train set and evaluated using the test set of CrowdPose with TTA. Table~\ref{tab:crowdpose} summarizes the results compared with the state-of-the-art approaches.

Higher HRNet~\cite{Cheng_2020_CVPR} achieves the best accuracy on CrowdPose. With the same backbone, PoseDet achieves higher inference speed and competitive accuracy than Higher HRNet, showing the robustness of PoseDet. Further, The accuracy of PoseDet suppress top-down methods (Mask-RCNN~\cite{he2017mask}, AlphaPose~\cite{fang2017rmpe} and Simple Baseline~\cite{xiao2018simple}) by a large margin, especially in the heavily occluded cases ($\text{AP}^{\text{H}}$). The main reason is that these top-down methods use a person detector to separate person instances with rectangular boxes. As discussed in~\ref{sec:introduction}, bounding boxes perform poorly for occluded cases.

\begin{figure*}
\centering
\includegraphics[scale=0.265]{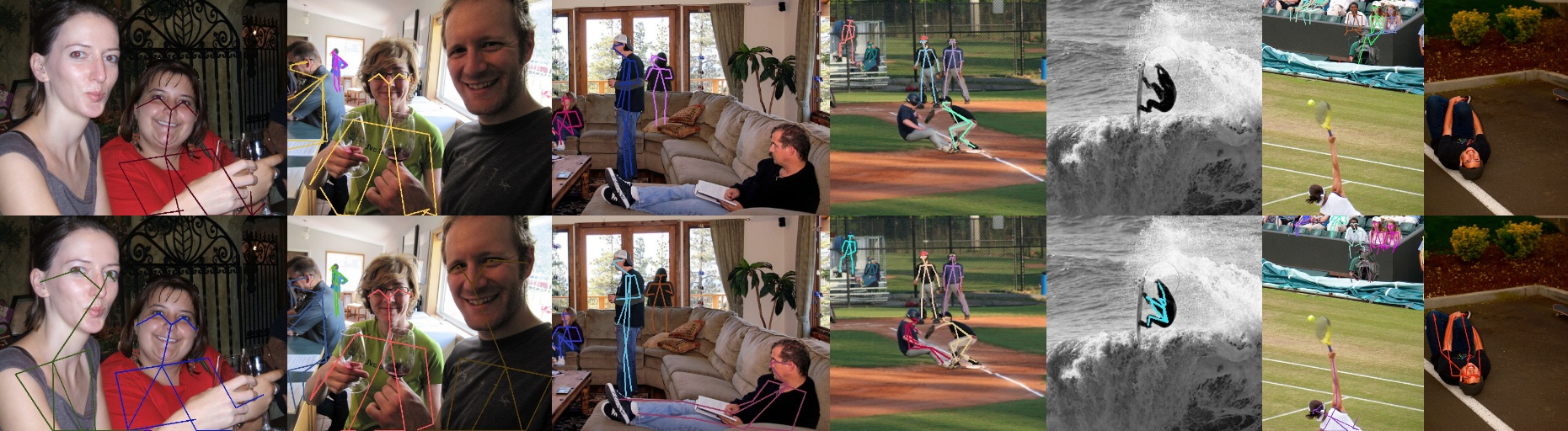}
\caption{Comparison of Point Anchors and PoseDet. The upper row is predictions of Point Anchors, and the lower row is results of PoseDet. PoseDet shows more robustness in corner cases (\ie, larger persons and rare poses).}
\label{fig:comparison1}
\end{figure*}

The reason for robustness is that PoseDet adopts keypoint-aware pose embedding. The offsets of pose embedding are learnable and supervised by keypoints. As shown in Figure~\ref{fig:comparison1}, PoseDet achieves a more accurate prediction than Point Anchors~\cite{wei2020point} in corner cases, where Point Anchors use fixed anchors as offsets.

\begin{table}
\centering
\scalebox{0.85}{
    \begin{tabular}{c|rrrrrr}
    \toprule
    JR  & $\text{AP}$ & $\text{AP}^{\text{50}}$ & $\text{AP}^{\text{75}}$ & $\text{AP}^{\text{E}}$ & $\text{AP}^{\text{M}}$ & $\text{AP}^{\text{H}}$ \\
    \midrule
     & 51.7 & 85.2 & 54.9 & 57.8 & 52.1 & 45.5\\
    \checkmark & 65.9 & 87.7 & 76.1 & 73.6 & 66.8 & 56.9\\

    \bottomrule
    \end{tabular}
    }
\caption{Effect of joint refinement (JR) on CrowdPose with backbone HRNet-W48.}
\label{tab:ablation2}
\end{table}

\begin{table}
\centering
\setlength\cellspacetoplimit{3pt}
\setlength\cellspacebottomlimit{0pt}
\scalebox{0.85}{
    \begin{tabular}{Sc|Sc|Sc|SrSrSr}
    \toprule
     Centroid type & NMS type & IS  & $\text{AP}^{\text{E}}$ & $\text{AP}^{\text{M}}$ & $\text{AP}^{\text{H}}$ \\
    \hline
     \multirow{4}*{Bounding box} & \multirow{2}*{IoU} & & 72.4 & 65.3 & 54.4 \\
    \cline{3-6}
    
      &  &\checkmark & 73.3 & 65.9 & 55.2 \\
    \cline{2-6}
    
      & \multirow{2}*{OKS} & & 72.9 & 66.1 & 55.7 \\ 
    \cline{3-6}
    
      &  & \checkmark & 73.6 & 66.7 & 56.6\\
    \cline{1-6}
    
     \multirow{4}*{Keypoints} & \multirow{2}*{IoU}& & 72.6 & 65.5 & 55.0  \\
    \cline{3-6}
    
      &  & \checkmark & 73.3 & 66.1 & 55.7\\
    \cline{2-6}
    
      & \multirow{2}*{OKS}& &  72.9      & 66.2      & 56.6   \\ 
    \cline{3-6}
    
      &  & \checkmark & \textbf{73.6}      & \textbf{66.8}      & \textbf{56.9}\\
    % \hline
    \bottomrule
    \end{tabular}
    }
\caption{Effect of keypoint-based operations on CrowdPose with backbone HRNet-W48. (\textit{IS: Intermediate supervision})}
\label{tab:ablation}
\end{table}

\begin{figure*}
\centering
\includegraphics[scale=0.301]{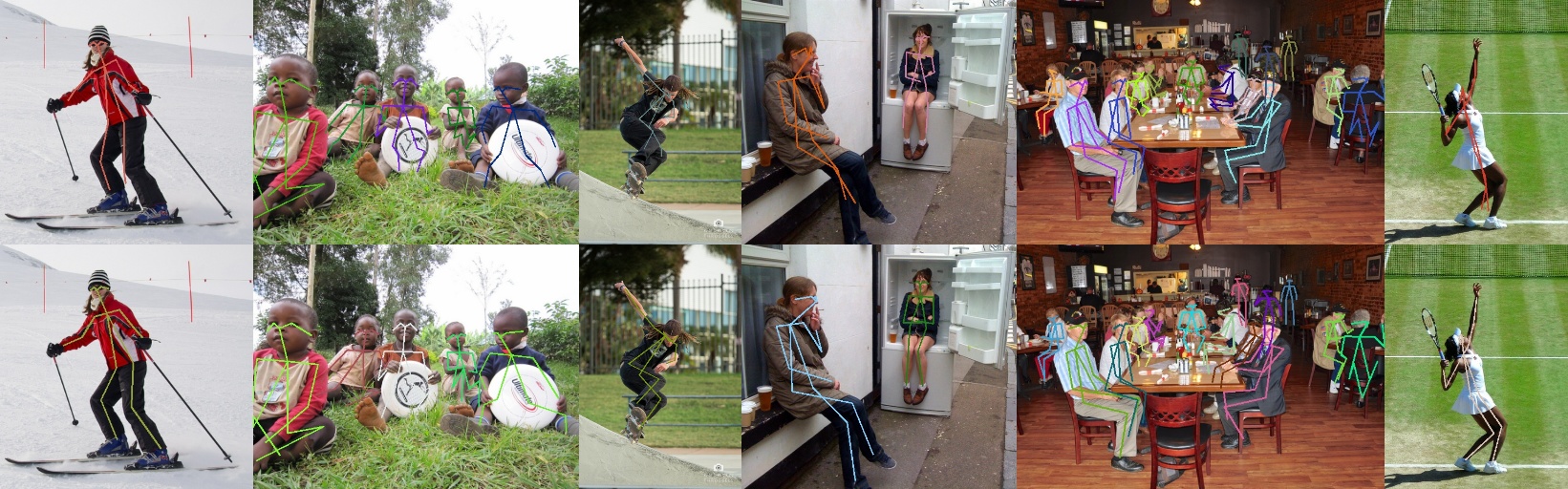}
\caption{Predictions of PoseDet, where the upper row is the coarse predictions and the lower row is the finer predictions.}
\label{fig:comparison2}
\end{figure*}

\subsection{Ablation study}
\label{sec:ablation}

PoseDet is basically a detector. Displacement refinement and keypoint-based operations are adopted to make it friendly for pose estimation. We perform ablation experiments to analyze the effect of these strategies. All experiments are conducted using PoseDet with backbone HRNet-W48. Models are trained using the train set, tested on the test set of CrowdPose without TTA.

\textbf{Effect of displacement refinement.} Table~\ref{tab:ablation2} shows the results of using displacement refinement (DR). PoseDet conducts DR via pose embedding based on the coarse prediction. The metrics, $\text{AP}^{\text{50}}$ and $\text{AP}^{\text{75}}$, measure predictions with OKS higher than 50 and 75 respectively.  We can find out that PoseDet without DR already achieves strong performance on the looser metric (85.2 $\text{AP}^{\text{50}}$). DR improve the accurate prediction by a large margin (+21.2 $\text{AP}^{\text{75}}$). The comparison demonstrates the effectiveness of pose embedding on encoding the bias of the coarse prediction.

The qualitative analysis in Figure~\ref{fig:comparison2} comes to consistent results. In the upper row, coarse predictions cover most keypoints. A high $\text{AP}^{\text{50}}$ is achieved with the coarse prediction. It also provides offsets for the DCN layer to generate keypoint-aware pose embedding, where the offsets are closely relative to keypoints. In the upper row, some keypoints that are not accurate enough are refined by predictions of pose embeddings. Estimation of poses is handled in a coarse-to-fine manner successfully.

\textbf{Effect of keypoint-based operations.} We compare PoseDet with different strategies of target assignment and NMS, as well as intermediate supervision in Table~\ref{tab:ablation}. 

The centroid of the bounding box is widely used as a basis to assign targets in object detection~\cite{qiu2020borderdet,tian2019fcos}. Pose estimation contains information on keypoints, which is more informative than the bounding box. We make use of it by replacing the centroid of the bounding box with the centroid of keypoints. We argue that the centroid of keypoints works better in occluded cases. We compare the performance that assigning a GT pose to the closest candidate according to the centroid of the bounding box or keypoints during training. In the results, we can find out that the centroid of keypoints improves the accuracy consistently.

As discussed in Section~\ref{sec:nms}, OKS is a more discriminate measurement to distinguish persons in the crowd than IoU. We conduct experiments with IoU NMS and OKS NMS. OKS NMS improves the performance of PoseDet in the crowd scenes (+1.4 $\text{AP}^\text{H}$ on average). 

The keypoints are further used in intermediate supervision. We use GT keypoints to generate Gaussian maps as supervision of FPN, forcing the FPN to perceive body joints. The accuracy is improved consistently.

\section{Conclusion and Future Works}
In this paper, we present keypoint-aware pose embedding, which can be seen as a general strategy to efficiently extract and aggregate informative feature representation. 
We propose PoseDet, a simple yet efficient method for multi-person pose estimation, in conjunction with the pose embedding.  
PoseDet simplifies the procedures of estimating, predicting accurate poses at an unprecedented speed.
We conduct displacement refinement and keypoint-based operations to improve the accuracy of PoseDet.

The future works include the applications to other instance-level recognition tasks, \eg , person re-identification, pose tracking, as well as the investigation on selecting the number of keypoints adaptively.

\bibliographystyle{named}
\bibliography{main}
\end{document}